\documentclass[runningheads]{llncs}

\usepackage{eccv}

\usepackage{eccvabbrv}

\usepackage{graphicx}
\usepackage{booktabs}

\usepackage[accsupp]{axessibility}  

\usepackage{algorithm}
\usepackage{algorithmic}
\usepackage{amsmath}
\usepackage{bm}
\usepackage{amssymb} 
\usepackage{xcolor}
\usepackage{booktabs}
\usepackage{multirow}
\usepackage{makecell}
\usepackage{colortbl}      
\usepackage{bbding}
\usepackage{placeins}
\usepackage{adjustbox}  
\usepackage{caption}
\usepackage{soul}
\usepackage{pifont} 

\usepackage{hyperref}

\usepackage{orcidlink}

\begin{document}

\title{Boosting Text-Driven Video Segmentation via Geometry-Aware Distillation} 

\titlerunning{GeoLaV}

\author{Tianyu Zhu\inst{1}\orcidlink{0009-0005-9540-1353}\textsuperscript{*} \and
Yingping Liang\inst{1}\orcidlink{0000-0001-5385-0015}\textsuperscript{*} \and
Hesong Li\inst{1}\orcidlink{0009-0002-0538-400X} \and
Ying Fu\inst{1}\orcidlink{0000-0002-6677-694X}\textsuperscript{\dag}}

\authorrunning{T.~Zhu et al.}

\institute{Beijing Institute of Technology, Beijing, China\\
\email{\{zhutianyu,liangyingping,lihesong2,fuying\}@bit.edu.cn}}

\maketitle

\makeatletter
\begingroup
\renewcommand{\thefootnote}{\fnsymbol{footnote}}
\footnotetext[1]{Equal contribution. \quad \textsuperscript{\dag} Corresponding author.}
\endgroup
\makeatother

\begin{abstract}
  Text-driven Referring Video Object Segmentation (RVOS) aims to locate and segment target objects in videos given natural language. However, existing models are typically trained on 2D image or video datasets with naive segmentation losses, which overlooks the geometric consistency across frames and leads to weak spatial understanding.
  In this paper, we propose \textbf{Geo}metry-enhanced \textbf{La}nguage-guided \textbf{V}ideo segmentation (\textbf{GeoLaV}), a two-stage framework that distills 3D geometric knowledge from images to enhance text-driven video segmentation. In the first stage, we perform monocular geometry pretraining with monocular novel-view synthesis, enabling the model to acquire geometry-consistent visual representations via spatial alignment on large-scale single-image datasets. In the second stage, we introduce geometry-aware distillation and fine-tune the model on video segmentation datasets, transferring 3D structural knowledge from a general 3D prior model. This process reinforces 3D awareness and improves both spatiotemporal coherence and language grounding in segmentation.
  Extensive experiments show that our method using only image segmentation data already provides notable zero-shot generalization in RVOS. When combined with geometry-aware distillation for fine-tuning on videos, our method achieves state-of-the-art performance across multiple RVOS benchmarks. The code is available at \url{https://github.com/Tony1882880/GeoLaV}.
\end{abstract}

\section{Introduction}
\label{sec:intro}

Segmenting objects in videos from natural language descriptions is a critical task in multimodal video understanding \cite{vid_understand_1,vid_understand_2}. Text-driven Referring Video Object Segmentation (RVOS) \cite{rvos_1,DsHmp} aligns visual dynamics with linguistic semantics to localize the described target. Traditional video segmentation methods \cite{trad_vid_seg_1,trad_vid_seg_2}, such as mask propagation-based \cite{trad_vid_seg_3} or category-specific approaches \cite{trad_vid_seg_5}, rely on predefined labels and lack flexibility in open-world scenarios. Recently, RVOS enables free-form language to specify objects by motion and appearance, supporting diverse applications \cite{vid_app_1,vid_app_2}. Advances in vision-language models have further boosted this paradigm, driving progress in multimodal understanding.

\begin{figure}
    \centering
    \includegraphics[width=1.0\linewidth]{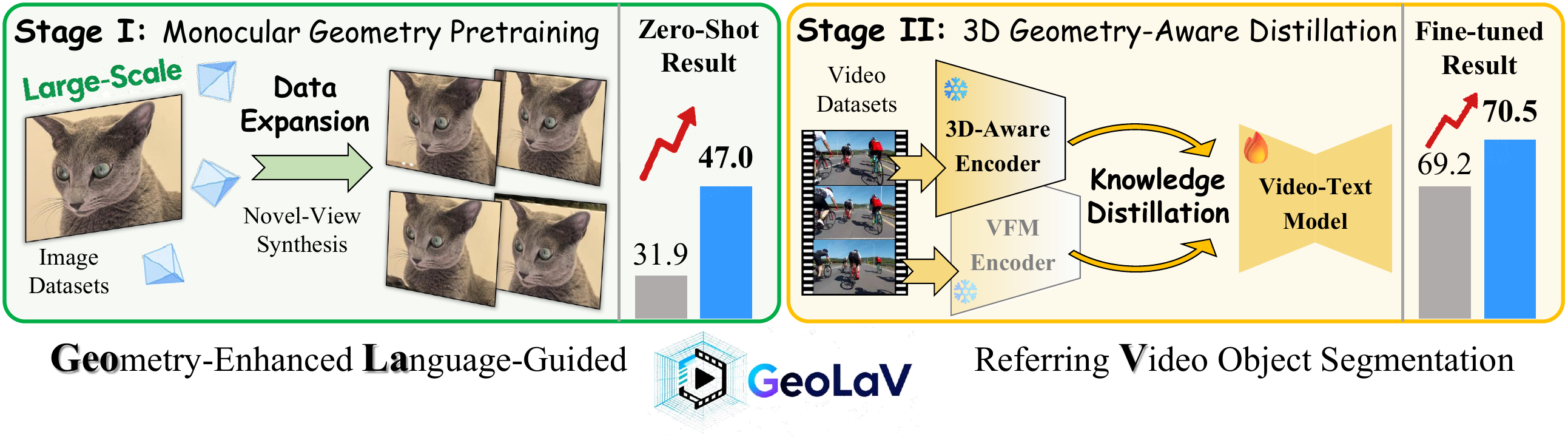}
    \caption{\textbf{Geo}metry-enhanced \textbf{La}nguage-guided referring \textbf{V}ideo object segmentation (GeoLaV) adopts a two-stage paradigm: \textbf{Stage I} performs monocular geometry pretraining via novel-view synthesis to learn geometry-consistent features; \textbf{Stage II} applies geometry-aware distillation with 3D teachers for spatiotemporal understanding.}
    \label{fig:teaser}
\end{figure}

However, the pretraining stage of most existing video segmentation methods \cite{SAMWISE,CLIP-VIS,OVFormer} remains largely dependent on large-scale 2D image datasets such as COCO \cite{COCO}, due to the scarcity and high cost of video annotations \cite{MeViS,Ref-Youtube-VOS}. These image-based pretraining datasets lack cross-frame geometric consistency and viewpoint variations, which are important for learning spatial geometry across views. As a result, models exhibit limited geometric awareness, making it difficult to maintain consistent object representations under viewpoint changes.

A further limitation arises from the fine-tuning paradigm of current video segmentation methods \cite{SAM2,SAMWISE}, which typically depend on memory banks or temporal attention to implicitly model cross-frame relations. However, they are often trained with naive segmentation losses on datasets of 2D images, overlooking 3D geometric consistency across frames and weakening spatial understanding. Without knowledge of 3D geometry, these models struggle to maintain stable tracking and consistent segmentation under complex camera motion and large object displacement, hindering the accurate modeling of dynamic scenes.

In this paper, we propose a novel two-stage framework, \textbf{Geo}metry-enhanced \textbf{La}nguage-guided \textbf{V}ideo segmentation (\textbf{GeoLaV}), which addresses these limitations by explicitly integrating geometric priors into text-driven video segmentation. Specifically, \textbf{in the first stage}, we introduce monocular geometry pretraining, where multiple novel-view images are synthesized from a single image under controlled camera motion to form geometry-consistent view sequences in an efficient and scalable manner. Rather than modeling complex temporal dynamics, this stage enforces cross-view structural consistency and encourages viewpoint-robust feature learning from synthetic data. With the aid of a visual foundation model encoder (\textit{e.g.}, DINOv2~\cite{DINOv2}, DINOv3~\cite{DINOv3}), the synthesized views are supervised in a distillation-style manner, encouraging the model to learn geometry-consistent representations across views and providing a geometry-aware initialization for subsequent fine-tuning on real video datasets.

Furthermore, \textbf{in the second stage}, we perform 3D geometry-consistent distillation to further enhance the model's understanding of spatial structure and motion. We incorporate general 3D models (\textit{e.g.}, VGGT \cite{VGGT}, ${\pi}^3$ \cite{pi3}) as teachers to provide 3D geometry priors and transfer geometry-aware representations into the segmentation framework. Through this distillation process, the model learns to maintain the geometric consistency of objects across frames, improving its capability to track targets under complex motions and viewpoint changes. This stage complements the monocular pretraining and enables our GeoLaV to jointly leverage semantic understanding and 3D geometric priors for text-driven video segmentation. Experiments demonstrate the state-of-the-art performance of our method across benchmarks. 

In summary, our main contributions are as follows:
\begin{itemize}
\item We propose \textbf{GeoLaV}, a two-stage framework that integrates single-view novel-view synthesis and 3D geometric priors to learn cross-frame consistency for text-driven video segmentation.
\item In \textbf{Stage I}, we perform monocular geometry pretraining via novel-view synthesis to learn geometry-consistent and temporally coherent features from large-scale 2D data.
\item In \textbf{Stage II}, we introduce 3D geometry-aware distillation to transfer 3D structural priors from general 3D models, enhancing spatial and temporal coherence across frames.
\end{itemize}

\section{Related Work}
\label{sec:related}

\noindent\textbf{Text-Driven Referring Video Segmentation.}
Text-driven RVOS \cite{TCE-RVOS,ReferFormer,SOC,DsHmp,guchangyu,SAMWISE} aims to segment target objects in videos given natural language descriptions. Early methods relied on two-stream architectures combining visual and textual features, later evolving into transformer-based frameworks that enable end-to-end multimodal learning \cite{MTTR,URVOS}. MTTR \cite{MTTR} first unified vision-language reasoning in a single Transformer, achieving strong results on A2D-Sentences \cite{A2D-Sentences} and Ref-YouTube-VOS \cite{Ref-Youtube-VOS}. SOC \cite{SOC} introduced semantic-assisted object clustering to enhance inter-frame consistency, while MUTR \cite{MUTR} employed a unified temporal transformer for long-range reasoning. To handle motion-sensitive expressions, DsHmp \cite{DsHmp} decoupled static and motion perception for finer temporal grounding. Recent lightweight frameworks such as SAMWISE \cite{SAMWISE} further leverage strong segmentation priors from SAM2 \cite{SAM2} through temporal adapters and memory-guided fusion, achieving better performance under a compact design.

With the development of deep learning \cite{qwen3vl,CJE1,CJE2,CJE3,3D-B2U,SFIN,SFIN2,FCDFusion,lhs-small,Interfaces}, large vision-language models (Large VLMs) have been explored for text-driven RVOS but remain computationally demanding. LISA \cite{LISA} employed a large multimodal LLM to reason about segmentation masks from natural prompts, VISA \cite{VISA} extended such reasoning to video through a mask decoder, and GLUS \cite{GLUS} unified global-local reasoning within a single model. Despite their impressive performance, these Large VLM-based methods rely heavily on large-scale pretraining and contain enormous parameters and computation costs, leading to slow inference. In contrast, non-Large VLM-based methods are more efficient but often overlook geometric cues, limiting spatial consistency across frames.

\begin{figure*}
  \begin{center}
  \includegraphics[width=1.0\linewidth]{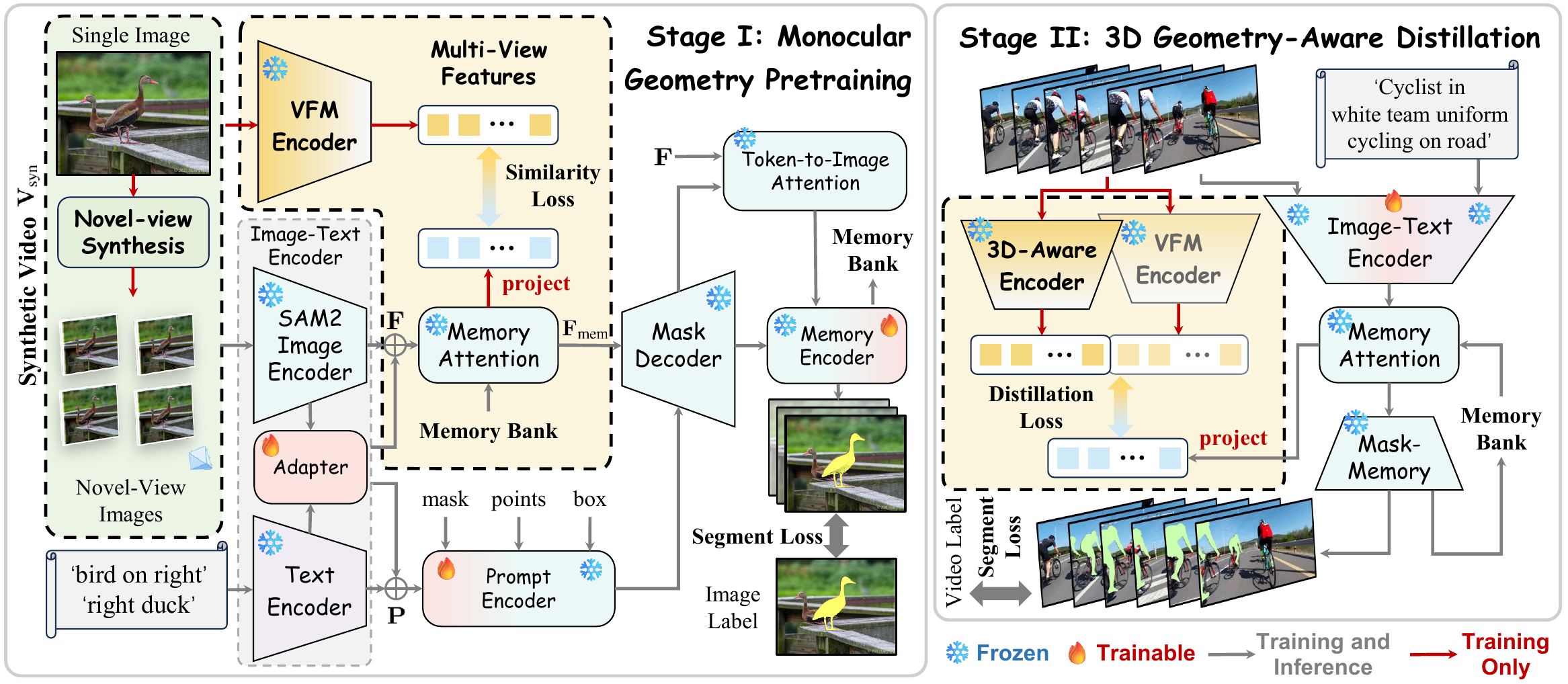}
  \end{center}
\caption{\textbf{Overview of our GeoLaV.} In the first stage, the monocular geometry pretraining process synthesizes novel views from single images and employs a visual foundation model encoder (\textit{e.g.}, DINOv3 \cite{DINOv3}) to learn inter-frame semantic representations. In the second stage, the geometry-aware distillation process utilizes a frozen 3D-aware teacher (\textit{e.g.}, $\pi^3$ \cite{pi3}) to inject 3D geometric priors into the video model, enhancing its spatiotemporal segmentation capability. Stage II shares the same backbone architecture as Stage I (illustrated in a simplified form in the figure) and is initialized with the pretrained weights from Stage I.}
\label{fig:framework}
\end{figure*}

\noindent\textbf{3D-Aware Geometry Representation Learning.}
Recent progress in 3D geometry representation learning \cite{flowanything,yuran,liangyingping,liangyingping2,TriMSOD} has advanced from monocular depth estimation to large-scale 3D-aware foundation models. Transformer-based approaches, such as DPT \cite{DPT}, Depth Anything v2 \cite{depthanything,depthanythingv2}, and Marigold \cite{marigold}, demonstrate that large-scale monocular supervision from synthetic and pseudo-labeled images enables encoders to learn robust depth priors from single images. Meanwhile, works such as 3D-to-2D distillation \cite{3Dto2D} show that geometric knowledge from 3D representations can be effectively transferred to 2D perception tasks. Building on this paradigm, recent 3D foundation models, including MASt3R \cite{MASt3R}, OpenScene \cite{OpenScene}, VGGT \cite{VGGT}, and $\pi^3$ \cite{pi3}, further extend geometric reasoning toward multi-view consistency and open-vocabulary 3D understanding by jointly modeling depth, camera pose, and semantic features.

Moreover, recent findings \cite{3DRS} reveal that multimodal large language models lack intrinsic 3D awareness and benefit significantly from geometry-informed supervision. These insights motivate integrating 3D priors into text-driven video segmentation to enhance spatial coherence and temporal understanding.

\section{Method}
\label{sec:method}

In this section, we first outline the formulation and motivation of our \textbf{GeoLaV}. As shown in \cref{fig:framework}, we introduce the Monocular Geometry Pretraining (MGP) stage to capture cross-view geometric consistency from synthesized videos, and the Geometry-Aware Distillation (GAD) stage to inject 3D geometric priors into text-driven referring video object segmentation.

\subsection{Formulation and Motivation}
Given a video sequence $\mathbf{V}=\{\mathbf{I}_t\}_{t=1}^{T}$, where each frame $\mathbf{I}_t$ denotes an RGB image, and a referring sentence $\mathbf{W}=\{w_i\}_{i=1}^{L}$ describing the target object, text-driven referring video segmentation aims to predict a sequence of binary masks $\mathbf{M}=\{\mathbf{M}_t\}_{t=1}^{T}$ indicating the object regions in all frames. 
Formally, the task can be formulated as:
\begin{equation}
\mathbf{M} = \mathcal{G}(\mathbf{V}, \mathbf{W}),
\end{equation}
where $\mathcal{G}$ denotes the segmentation function integrating visual features from $\mathbf{V}$ and linguistic semantics from $\mathbf{W}$ to produce coherent masks across frames.

However, there are still two main challenges. First, existing RVOS methods are typically pretrained on RefCOCO/+/g \cite{RefCOCO,RefCOCOg}, which consist solely of static images. Such pretraining provides only spatial correspondence between text and individual frames but fails to expose the model to cross-view geometric variation. To address this, we synthesize geometry-consistent multi-view images from a single frame to form pseudo videos, allowing the model to perceive continuous visual transitions. We then introduce a visual foundation model (VFM) encoder to learn cross-frame semantic relationships within these generated sequences, enabling the model to build a more robust spatio-temporal understanding before fine-tuning on real videos.

Second, existing frameworks lack explicit modeling of 3D geometric consistency between frames, which is crucial for accurately tracking moving objects in dynamic scenes. To overcome this, we incorporate a 3D-aware encoder that learns frame-to-frame geometric relationships by aligning with features from pretrained 3D perception models. This higher-dimensional geometric supervision enables the model to capture object structure and motion beyond 2D appearance cues, resulting in more stable and geometry-consistent video segmentation.

\begin{figure*}[t]
    \centering
    \includegraphics[width=\linewidth]{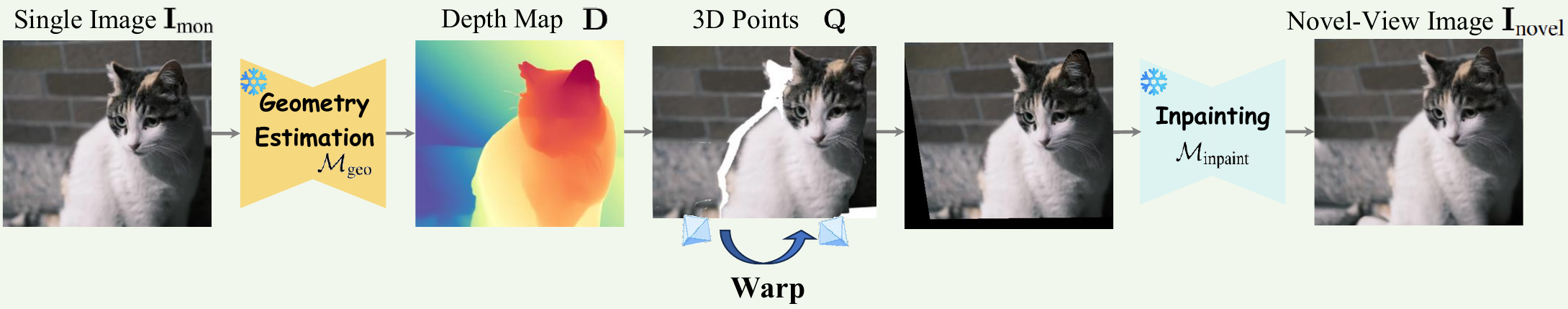}

    \caption{\textbf{Pipeline of novel-view image synthesis.} 
    Starting from a single RGB input, we estimate depth using a feed-forward geometry model, project pixels into 3D space, and apply virtual camera transformations to obtain new viewpoints. An occlusion mask and an inpainting network are used to refine the rendered views, ensuring geometric coherence and visual completeness.}
    \label{fig:novel_pipeline}

\end{figure*}

\subsection{Monocular Geometry Pretraining}
\label{sec:mgp}
\noindent\textbf{Overall Architecture.}
Given the large-scale availability of image datasets, most pretraining frameworks rely solely on static 2D images, which overlook the temporal relationships crucial for video understanding. To bridge this gap, we introduce a monocular geometry pretraining framework that explicitly learns inter-frame consistency by converting single images into geometry-consistent synthetic videos. Each synthetic video \( \mathbf{V}_{\text{syn}} = \{\mathbf{I}^{t}\}_{t=1}^{T} \) is constructed by combining the original image with its 3D geometry-consistent novel-view renderings, generated through the synthesis strategy described later. These videos serve as training inputs, allowing the model to capture semantics across frames while leveraging large-scale image-based data.

As shown in Figure \ref{fig:framework}, the video \(\mathbf{V}_{\text{syn}}\) is processed by the frozen SAM2 \cite{SAM2} image encoder $\mathcal{E}_{\text{img}}$, while the textual description \( \mathbf{W} \) is passed through the frozen text encoder $\mathcal{E}_{\text{txt}}$. Both encoders are hierarchical with progressive downsampling. Let \(l\) denote the encoder level and \(t\) the frame index; the per-level visual and textual features are denoted as
\begin{eqnarray}
\{\mathbf{F}^{t,l}\}_{l=1}^{L} = \mathcal{E}_{\text{img}}(\mathbf{I}^{t}), 
\qquad 
\{\mathbf{P}^{l}\}_{l=1}^{L} = \mathcal{E}_{\text{txt}}(\mathbf{W}).
\end{eqnarray}
A trainable cross-modal adapter $\mathcal{A}_{\theta}$ is inserted at each level to align the two modalities in a shared representation space, while keeping both encoders frozen:
\begin{eqnarray}
\mathbf{F}^{\prime\,t,l} = \mathcal{A}^{l}_{\theta}\!\big(\mathbf{F}^{t,l},\, \mathbf{P}^{l}\big), 
\quad l=1,\dots,L,\ \ t=1,\dots,T.
\end{eqnarray}

The aligned features are passed through the frozen memory attention module to obtain memory-enhanced representations \( \mathbf{F}^{t,l}_{\text{mem}} \). The prompt encoder combines frozen SAM2 \cite{SAM2} weights with a small set of trainable parameters to encode task prompts. The outputs of the prompt encoder and memory attention are fed into the mask decoder, whose predictions are refined by the memory encoder to produce segmentation masks. Meanwhile, the memory encoder continuously updates a memory bank, which stores temporal information and is reused by the memory attention module in subsequent frames.

In parallel, the synthesized video is also processed by a visual foundation model (VFM) encoder, such as DINOv3 \cite{DINOv3}, to extract features that provide semantic supervision. Since the alignment is applied before prompt-conditioned decoding, it operates on global frame-level representations rather than object-specific features. The overall training objective is formulated as
\begin{eqnarray}
\mathcal{L}_{\text{total}} = \mathcal{L}_{\text{seg}} + \mathcal{L}_{\text{sim}},
\end{eqnarray}
where segmentation loss \( \mathcal{L}_{\text{seg}} = \mathcal{L}_{\text{dice}} + \mathcal{L}_{\text{mask}} \) combines Dice and mask losses following previous works, and similarity loss \( \mathcal{L}_{\text{sim}} \) aligns the VFM features with the projected memory-attention features (detailed later). This design enables the model to jointly learn spatial semantics and temporal consistency from large-scale image data without requiring real video annotations.

\noindent\textbf{Novel-View Images Synthesis.}
To preserve the 3D geometric fidelity of the original image during synthetic view-sequence generation, we design a geometry-consistent multi-view synthesis framework, as illustrated in \cref{fig:novel_pipeline}. Inspired by L2M~\cite{L2M}, we employ recent feed-forward 3D geometry estimation models (\textit{e.g.}, $\pi^3$~\cite{pi3}, VGGT~\cite{VGGT}) that directly infer scene geometry from a monocular RGB image. These models predict scale-invariant depth or point maps in a reference-free manner, providing a reliable geometric foundation for synthesizing novel views.

\begin{figure}[t]
    \centering
    \includegraphics[width=0.88\linewidth]{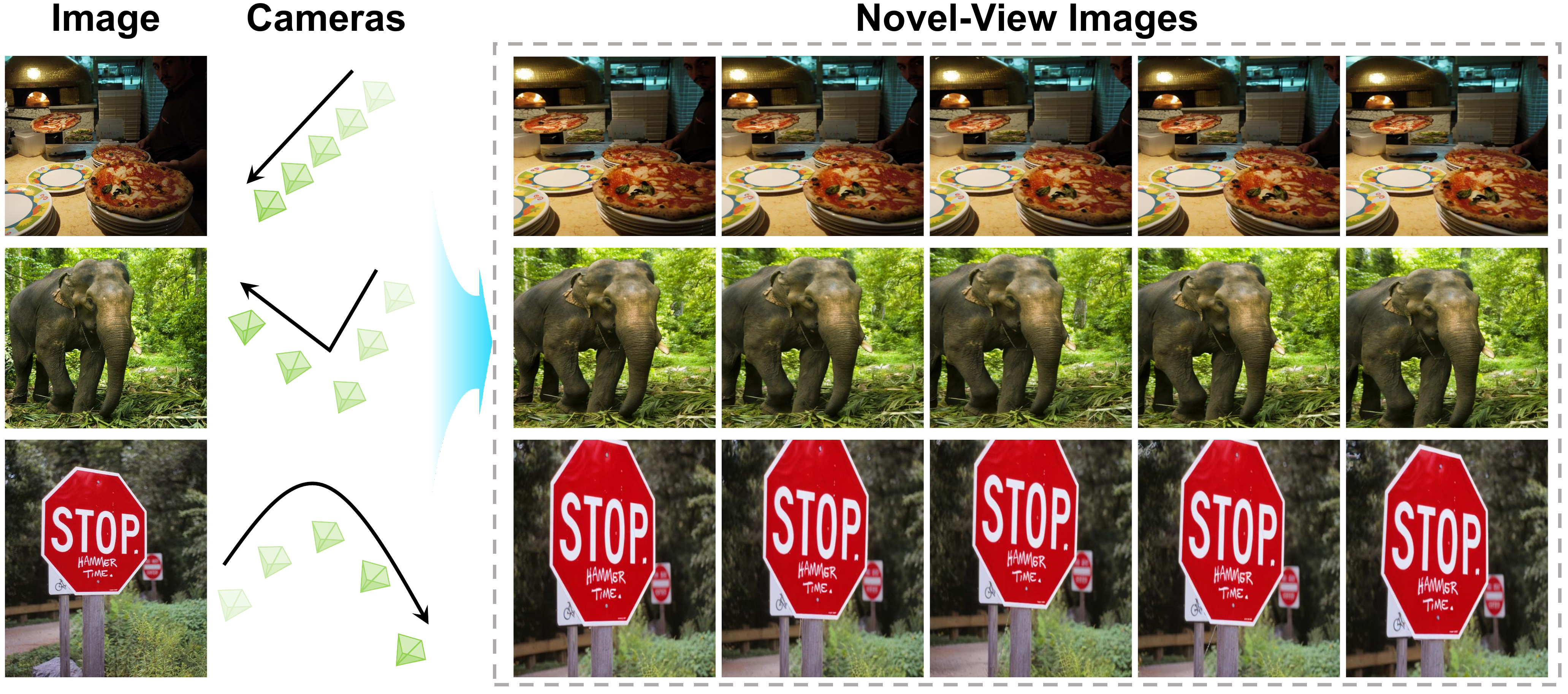}
    \setlength{\abovecaptionskip}{0mm}
    \caption{\textbf{Examples of synthesized novel-view sequences for our Monocular Geometry Pretraining (MGP).} Each row corresponds to a continuous camera trajectory (\textit{e.g.}, linear, piecewise-linear, and curved motion). Adjacent frames are generated through physically valid rigid transformations in 3D space, preserving consistent geometric structure and depth–camera correspondence across views.}
    \label{fig:novel_examples}

\end{figure}

Given a monocular image \( \mathbf{I}_{\text{mon}} \), the network estimates a dense depth map \( \mathbf{D} \). To simulate the inherent scale ambiguity in monocular reconstruction, we introduce a random scaling factor \( a \) and a bias term \( b \), \textit{i.e.},
\begin{eqnarray}
    \mathbf{D} = a \mathcal{M}_{\text{geo}}(\mathbf{I}_{\text{mon}}) + b,
\end{eqnarray}
where \( \mathcal{M}_{\text{geo}} \) denotes the chosen 3D geometry estimation model. Although the predicted depth is not metrically precise, it preserves consistent 3D geometry with the original monocular image, ensuring that synthesized frames remain geometrically aligned with the source view.

Using the predicted depth, the image is lifted into 3D space by projecting each pixel \( (u,v) \) into 3D coordinates through a sampled intrinsic matrix \( \mathbf{K} \), forming a dense and geometry-consistent point set \( \mathbf{Q} = \{(X,Y,Z)\} \). As shown in \cref{fig:novel_examples}, instead of sampling independent camera poses, we generate a short continuous camera trajectory across frames. The trajectory may follow linear, piecewise-linear, or curved motion patterns to increase diversity while preserving geometric consistency. Starting from an initial pose, subsequent poses are obtained via small incremental rigid transformations under a physically valid camera projection model, forming a coherent view sequence with real geometric correspondence between adjacent frames.

The transformed 3D points are then reprojected into the image plane to produce geometry-consistent novel-view renderings \( \mathbf{I}_{\text{novel}} \). To mitigate occlusion and disocclusion artifacts, an occlusion mask \( \mathbf{M} \) is computed and an inpainting network \( \mathcal{M}_{\text{inpaint}} \) is adopted to fill missing regions:
\begin{eqnarray}
\mathbf{I}_{\text{syn}} = \mathcal{M}_{\text{inpaint}}(\mathbf{I}_{\text{novel}}, \mathbf{M}),
\end{eqnarray}
where \( \mathcal{M}_{\text{inpaint}} \) restores invisible areas based on surrounding contextual cues. The generated view sequences maintain consistent structure, illumination, and depth–camera correspondence across frames, establishing a geometry-coherent foundation for large-scale pretraining.

Our geometry-based pipeline is deterministic and computationally efficient. A short multi-frame sequence can be synthesized within seconds, enabling scalable data expansion for geometry-aware pretraining on large image datasets.

\noindent\textbf{Feature Projection Module.}
As shown in the left part of \cref{fig:framework}, the features extracted from the SAM2 \cite{SAM2} memory attention branch and the VFM encoder (\textit{e.g.}, DINOv3 \cite{DINOv3}) differ in both dimensionality and distribution.  
To achieve consistent representation learning, we introduce a lightweight projection head that transforms the memory-attention features into the same embedding space as the VFM representations.  
This projection head is implemented as a two-layer MLP with GELU activation,
\begin{eqnarray}
\mathbf{F}_{\text{proj}} = \mathcal{P}_{\theta}(\mathbf{F}_{\text{mem}}) 
= \mathbf{W}_{2}\,\sigma(\mathbf{W}_{1}\mathbf{F}_{\text{mem}}),
\end{eqnarray}
where \( \mathcal{P}_{\theta} \) denotes the projection function parameterized by learnable weights \( \mathbf{W}_{1} \) and \( \mathbf{W}_{2} \), and \( \sigma(\cdot) \) represents the GELU activation.  
The output dimension of this projection is dynamically matched to the feature dimension of the selected VFM encoder, ensuring compatibility across different visual foundation backbones.  
This projection is used only during training and introduces no additional computation or parameters at inference.

To inject structural and geometry-aware priors from the 3D and VFM encoders into the SAM2 \cite{SAM2} backbone, we perform feature alignment at the memory-attention level during fine-tuning. Since this representation is computed before prompt-conditioned mask decoding, it tends to capture general feature rather than object-specific features tied to a particular textual query. We employ a cosine similarity loss between the projected feature \( \mathbf{F}_{\text{proj}} \) and the corresponding VFM feature \( \mathbf{F}_{\text{VFM}} \), \textit{i.e.},
\begin{eqnarray}
\mathcal{L}_{\text{sim}} 
= 1 - \text{Sim}(\mathbf{F}_{\text{proj}}, \mathbf{F}_{\text{VFM}}) 
= 1 - 
\frac{\mathbf{F}_{\text{proj}} \cdot \mathbf{F}_{\text{VFM}}}
{\|\mathbf{F}_{\text{proj}}\| \, \|\mathbf{F}_{\text{VFM}}\|},
\end{eqnarray}
where \( \mathcal{L}_{\text{sim}} \) denotes the cosine similarity loss,
\( \text{Sim}(\cdot) \) computes cosine similarity,
and \( \|\cdot\| \) represents the \(L_2\) norm. The alignment is therefore applied to backbone memory features prior to prompt-conditioned decoding, while the primary supervision remains the text-driven segmentation objective. The similarity loss serves as an auxiliary regularizer that injects geometry-aware structural cues while compensating for the lack of supervision in synthesized frames.

\subsection{Geometry-Aware Distillation}
Building upon the pretraining stage, this step focuses on enhancing 3D geometric understanding and temporal stability through geometry-aware distillation. As illustrated in the right part of \cref{fig:framework}, the model is fine-tuned on real video datasets so that it can perceive structural variations across frames and maintain consistent object representations.

Each video sequence \(\mathbf{V}_{\text{real}} = \{\mathbf{I}^{t}\}_{t=1}^{T}\)\ is paired with text descriptions and processed by a 3D-aware encoder, such as \(\pi^3\)~\cite{pi3}, together with a VFM encoder. The 3D-aware encoder provides higher-dimensional geometric priors that capture structural and depth relationships across frames, while the VFM encoder contributes semantic priors that describe object appearance and category. By combining complementary cues, the model learns to perceive objects with stronger spatial coherence and achieves more stable tracking in dynamic scenes.

Similar to the monocular geometry pretraining stage, two projection heads are introduced to align the feature spaces of the encoders with the memory representation. Each projection head is implemented as a two-layer MLP with GELU activation,
\begin{equation}
\begin{aligned}
\mathbf{F}^{(k)}_{\text{proj}} &= \mathcal{P}^{(k)}_{\theta}\!\big(\mathbf{F}_{\text{mem}}\big)
= \mathbf{W}^{(k)}_{2}\,\sigma\!\big(\mathbf{W}^{(k)}_{1}\mathbf{F}_{\text{mem}}\big), \quad
k \in \{\text{3D},\, \text{VFM}\}.
\end{aligned}
\end{equation}
All notations are consistent with those defined in \cref{sec:mgp}.
Their output dimensions are matched to the corresponding encoders, and both heads are used only during training without affecting inference.

The supervision is established by cosine similarity losses computed between the projected memory features and the outputs of each encoder, with the overall distillation objective defined as
\begin{equation}
\begin{aligned}
\mathcal{L}_{\text{distill}} 
&= \mathcal{L}_{\text{3D}} + \mathcal{L}_{\text{VFM}} 
= \sum_{k \in \{\text{3D},\, \text{VFM}\}} \big[1 - \text{Sim}~\!\big(\, \mathbf{F}^{(k)}_{\text{proj}},\, \mathbf{F}_{k} \,\big)\big],
\end{aligned}
\end{equation}
where the similarity function \( \text{Sim}(\cdot) \) follows the definition in \cref{sec:mgp}.  
Through this dual-branch distillation, the backbone representation is regularized to encode richer geometric structure across frames. The text-conditioned mask prediction remains guided by the segmentation objective,
ensuring that refer-object specificity is preserved while benefiting from geometry-aware cues.

\begin{table*}[t]
\scriptsize
\centering
\setlength{\tabcolsep}{1.25mm}
\renewcommand\arraystretch{1.1}
\definecolor{lightgray}{gray}{0.9}
\definecolor{lightblue}{RGB}{220, 240, 255}
\caption{Quantitative Comparison of our Non-Large VLM-based method with state-of-the-art RVOS approaches on Ref-Youtube-VOS \cite{Ref-Youtube-VOS}, Ref-DAVIS17 \cite{Ref-DAVIS17}, and MeViS \cite{MeViS}. For non-large VLM-based methods, the best results are shown in \textbf{Bold} and the second-best results are \underline{Underlined}. We also include large VLM-based methods for reference, where the best results are shown in \textbf{Bold}. $\dag$ indicates methods evaluated with SAM2 \cite{SAM2} integration via GroundingDINO \cite{GroundingDINO}.}
{
\begin{tabular}{l|c|ccc|ccc|ccc}
\hline
\multirow{2}{*}{Method} & \multirow{2}{*}{\makecell[l]{Total\\Params}} 
 & \multicolumn{3}{c|}{Ref-YouTube-VOS} & \multicolumn{3}{c|}{Ref-DAVIS17} & \multicolumn{3}{c}{MeViS} \\
\cline{3-11}
 & & $\mathcal{J}\&\mathcal{F}$ & $\mathcal{J}$ & $\mathcal{F}$ & $\mathcal{J}\&\mathcal{F}$ & $\mathcal{J}$ & $\mathcal{F}$ & $\mathcal{J}\&\mathcal{F}$ & $\mathcal{J}$ & $\mathcal{F}$ \\
\hline
\rowcolor{lightgray} \multicolumn{11}{l}{\textit{\textbf{Large VLM}-based}} \\
LISA \cite{LISA} & 7B & 53.9 & 53.4 & 54.3 & 64.8 & 62.2 & 67.3 & 37.2 & 35.1 & 39.4 \\
VISA-7B \cite{VISA} & 7B & 61.5 & 59.8 & 63.2 & \textbf{69.4} & \textbf{66.3} & \textbf{72.5} & 43.5 & 40.7 & 46.3 \\
VISA-13B \cite{VISA} & 13B & 63.0 & 61.4 & 64.7 & 67.0 & 73.8 & 70.4 & 44.5 & 41.8 & 47.1 \\
One-Token-Seg-All & 3.8B & 61.7 & 60.2 & 63.3 & 67.7 & 63.8 & 71.5 & 42.3 & 39.4 & 45.2 \\
VideoGLaMM & 3.8B & - & - & - & - & - & -  & 45.2 & 42.1 & 48.2\\
GLUS$^S$ & 7B & 66.6 & 65.0 & 68.3 & - & - & -  & 50.3 & 47.5 & 53.2\\
GLUS$^A$ \cite{GLUS} & 7B & \textbf{67.3} & \textbf{65.5} & \textbf{69.0} & - & - & -  & \textbf{51.3} & \textbf{48.5} & \textbf{54.2} \\

\hline
\rowcolor{lightgray} \multicolumn{11}{l}{\textit{\textbf{Non-Large VLM}-based}} \\

MTTR \cite{MTTR} & - & 55.3 & 54.0 & 56.6 & - & - & -  & 30.0 & 28.8 & 31.2\\
TCE-RVOS \cite{TCE-RVOS} & - & 59.6 & 58.3 & 60.8 & 59.4 & 56.5 & 62.4 & - & - & - \\
ReferFormer \cite{ReferFormer} & 237M & 62.9 & 61.3 & 64.6 & 61.1 & 58.1 & 64.1 & 31.0 & 29.8 & 32.2 \\
SOC \cite{SOC} & 220M & 66.0 & 64.1 & 67.9 & 64.2 & 61.0 & 67.4 & - & - & - \\
OnlineRefer \cite{OnlineRefer} & 232M & 63.5 & 61.6 & 65.5 & 64.8 & 61.6 & 67.7 & 32.3 & 31.5 & 33.1 \\
LMPM \cite{MeViS} & 195M & - & - & - & - & - & - & 37.2 & 34.2 & 40.2 \\
DsHmp \cite{DsHmp} & 339M & 67.1 & 65.0 & 69.1 & 64.9 & 61.7 & 68.1 & - & - & - \\
DsHmp \cite{DsHmp} & 272M & - & - & - & - & - & - & 46.4 & 43.0 & 49.8 \\
MUTR \cite{MUTR} & 250M & 67.5 & 65.4 & 69.6 & 66.4 & 62.8 & 70.0 & - & - & - \\
GroundingDINO \cite{GroundingDINO} $^\dag$ & 240M & 57.5 & 55.6 & 59.5 & 66.4 & 62.8 & 69.9 & 37.7 & 34.9 & 40.5 \\
LAVT \cite{LAVT} & 239M & 65.8 & 63.6 & 67.9 & - & - & - & - & - & - \\
SSA \cite{SSA} & 474M & 64.3 & 62.2 & 66.4 & 67.3 & 64.0 & 70.7 & 48.6 & 44.0 & 53.2 \\
SAMWISE \cite{SAMWISE} & 150M & 67.2 & 65.2 & 69.3 & 68.5 & 65.6 & 71.5 & 48.3 & 45.4 & 51.2 \\
SAMWISE \cite{SAMWISE} & 202M & 69.2 & \underline{67.8} & 70.6 & \underline{70.6} & \underline{67.4} & \underline{73.5} & \underline{49.5} & \underline{46.6} & 52.4 \\
ReferDINO \cite{ReferDINO} & 230M & \underline{69.3} & 67.0 & \underline{71.5} & 68.9 & 65.1 & 72.9 & 49.3 & 44.7 & \textbf{53.9} \\
\rowcolor{lightblue} \textbf{GeoLaV (Ours)} & 202M & \textbf{70.5} & \textbf{69.1} & \textbf{71.8} & \textbf{72.5} & \textbf{69.9} & \textbf{75.2} & \textbf{50.0} & \textbf{47.4} & \underline{52.9} \\
\hline
\end{tabular}}
\label{tab:main_result}

\end{table*}

\section{Experiments}
\label{sec:exp}

In this section, we evaluate our GeoLaV through comprehensive experiments. We first outline the experimental setup, followed by the main results on multiple referring video segmentation benchmarks. Finally, we discuss the effects of each design and the benefits introduced by geometry pretraining. Additional experiments and visualizations are provided in the supplementary material.

\subsection{Experiment Setup}
\noindent\textbf{Datasets.}
Following the previous works \cite{DsHmp,SAMWISE}, we evaluate our approach on three referring video segmentation benchmarks, \textit{i.e.}, Ref-Youtube-VOS \cite{Ref-Youtube-VOS}, Ref-DAVIS17 \cite{Ref-DAVIS17}, and MeViS \cite{MeViS}. Ref-Youtube-VOS extends the YouTube-VOS benchmark by adding textual descriptions, containing 3,978 high-resolution videos and about 15K language expressions that refer to target objects across time. Ref-DAVIS17 builds upon the DAVIS17 dataset, enriching 90 videos with over 1.5K linguistic annotations, supporting referring-expression segmentation in scenes with multiple moving objects and complex backgrounds. MeViS includes 2,006 videos and approximately 28K motion-centric expressions that capture complex motion patterns and emphasize motion description rather than static cues.

\noindent\textbf{Implementation Details.}
Our model architecture is basically built upon the SAMWISE framework \cite{SAMWISE}, with modifications to incorporate geometry-aware feature learning. We employ DINOv3-ViT-L \cite{DINOv3} as VFM encoder and $\pi^3$ \cite{pi3} as 3D-aware encoder. \textbf{In the first stage,} we synthesize five novel-view images from each COCO \cite{COCO} single-view input to form pseudo videos. Following previous works \cite{HTML,SOC,SAMWISE}, the pretraining is performed on RefCOCO/+/g \cite{RefCOCO,RefCOCOg} for 6 epochs, with an initial learning rate of $1\times10^{-4}$ decayed to $2\times10^{-5}$. \textbf{In the second stage,} we fine-tune on video data from Ref-Youtube-VOS \cite{Ref-Youtube-VOS} for 4 epochs with a learning rate of $2\times10^{-6}$. The trained model is evaluated on Ref-Youtube-VOS and Ref-DAVIS17 \cite{Ref-DAVIS17}, and for MeViS \cite{MeViS} we train for 2 epochs under the same protocol. All experiments use the Adam optimizer on 4 NVIDIA A100 GPUs (80GB GPU memory each).

\noindent\textbf{Evaluation Metrics.}
We adopt evaluation metrics, including region similarity (\(\mathcal{J}\)), contour accuracy (\(\mathcal{F}\)), and their mean score \((\mathcal{J}\ \&\ \mathcal{F})\). For MeViS \cite{MeViS} and Ref-Youtube-VOS \cite{Ref-Youtube-VOS}, evaluations are performed via the official challenge servers, while Ref-DAVIS17 \cite{Ref-DAVIS17} is assessed using the official evaluation code.

\subsection{Main Results}
We present the main results to comprehensively evaluate GeoLaV. We first introduce the compared methods from both non-large and large VLM-based RVOS approaches. Then, we compare the overall performance on three benchmarks, showing consistent gains in segmentation quality.

\begin{figure*}[t]
\centering
\setlength{\belowcaptionskip}{0mm}
\setlength{\abovecaptionskip}{2mm}
\includegraphics[width=0.92\linewidth]{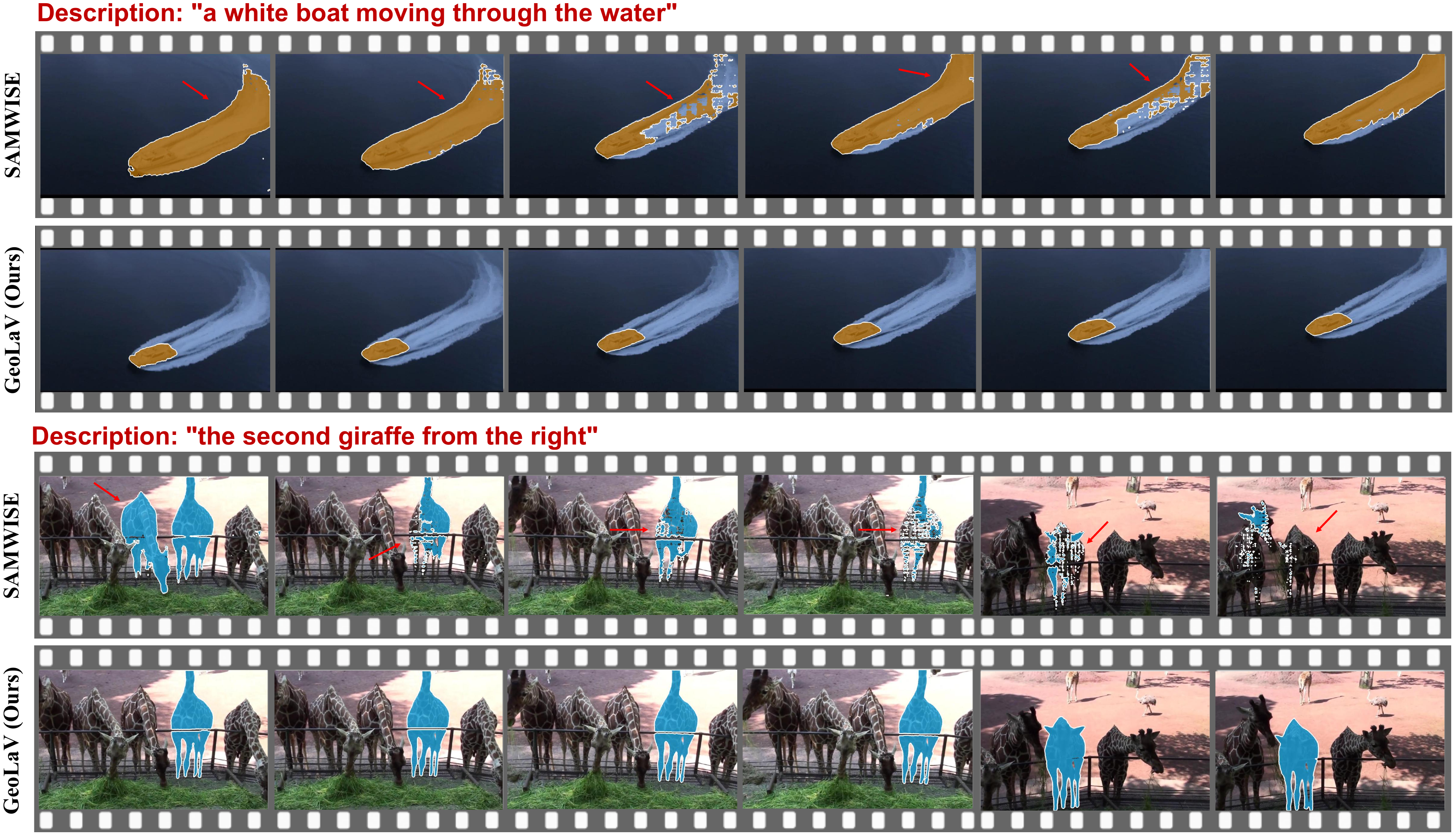}
\caption{Qualitative comparison with SAMWISE \cite{SAMWISE}. It represents the state-of-the-art method and serves as the base architecture of our model. Our GeoLaV more accurately identifies target objects and maintains stable tracking across challenging video frames.}
\label{fig:qualitative}
\end{figure*}

\noindent\textbf{Comparison Methods.}  
We compare GeoLaV with a wide range of RVOS approaches. The non-large VLM-based group includes MTTR \cite{MTTR}, TCE-RVOS \cite{TCE-RVOS}, ReferFormer \cite{ReferFormer}, SOC \cite{SOC}, OnlineRefer \cite{OnlineRefer}, LMPM \cite{MeViS}, DsHmp \cite{DsHmp}, MUTR \cite{MUTR}, GroundingDINO \cite{GroundingDINO}, SSA \cite{SSA}, SAMWISE \cite{SAMWISE}, and ReferDINO \cite{ReferDINO}, all having comparable parameter scales to our method. In addition, we include large VLM-based approaches such as LISA \cite{LISA}, VISA-7B/13B \cite{VISA}, One-Token-Seg-All \cite{O-T-S-A}, VideoGLaMM \cite{VideoGLaMM}, and GLUS$^{S/A}$ \cite{GLUS} as extra references for comparison on the same benchmarks. All reported results (except GroundingDINO) of these methods are directly taken from their original papers for fair comparison. For GroundingDINO, we report its performance when combined with SAM2~\cite{SAM2}, as it is originally designed for object detection.

\noindent\textbf{Quantitative Comparison of Full Framework.} 
As shown in \cref{tab:main_result}, GeoLaV achieves state-of-the-art performance across all three benchmarks. Among non-large VLM-based methods, our approach consistently ranks first on Ref-Youtube-VOS \cite{Ref-Youtube-VOS} (70.5 $\mathcal{J} \& \mathcal{F}$), Ref-DAVIS17 \cite{Ref-DAVIS17} (72.5 $\mathcal{J} \& \mathcal{F}$), and MeViS \cite{MeViS} (50.0 $\mathcal{J} \& \mathcal{F}$), surpassing the previous best results by 1.2, 1.9, and 0.5 points, respectively. Compared with recent strong competitors such as SAMWISE \cite{SAMWISE} and ReferDINO \cite{ReferDINO}, GeoLaV demonstrates notable improvements in both region similarity and contour accuracy, highlighting the benefits of incorporating geometric priors and temporally consistent reasoning. Furthermore, despite using a much smaller model size than large VLM-based approaches (\textit{e.g.}, GLUS$^A$ \cite{GLUS} with 7B parameters), GeoLaV achieves highly competitive results, demonstrating the efficiency and effectiveness of our framework.

\noindent\textbf{Qualitative Results of Full Framework.}
Since SAMWISE \cite{SAMWISE} represents the state-of-the-art and serves as the base architecture of our model, we conduct a direct visual comparison between the two. Additional visual comparisons are provided in the supplementary materials. Our GeoLaV produces more accurate and temporally stable segmentation across diverse scenes. In the first example of \cref{fig:qualitative}, SAMWISE mistakenly segments the white wake behind the boat as part of the target, while our method precisely captures only the boat region. In the second example, SAMWISE fails to consistently track the rightmost giraffe across frames, whereas our approach maintains stable and coherent masks over time. These results demonstrate that incorporating geometric priors effectively enhances spatial precision and temporal consistency, enabling fine-grained segmentation across complex video sequences.

\subsection{Discussions}
We analyze our GeoLaV from four aspects. We first evaluate the generalization of the geometry-consistent data expansion via zero-shot evaluation after pretraining, then examine the contributions of MGP and GAD. Finally, we analyze geometric accuracy and visualize learned features to understand distillatiWon.

\noindent\textbf{Zero-Shot Evaluation after Pretraining.}  
We perform zero-shot evaluation to validate the effectiveness of our geometry-consistent data expansion strategy in extending image pretraining to video segmentation. The model is trained only on RefCOCO/+/g \cite{RefCOCO,RefCOCOg} and directly tested on Ref-Youtube-VOS \cite{Ref-Youtube-VOS} and MeViS \cite{MeViS} without any video fine-tuning. For comparison, we use SAMWISE \cite{SAMWISE}, which represents the current state-of-the-art method and serves as the base architecture of our model. We further include other video-based methods that provide publicly available image-pretrained weights, such as LMPM \cite{MeViS}, DsHmp \cite{DsHmp}, and LAVT \cite{LAVT}, as well as referring image segmentation methods including DMMI \cite{DMMI}, ReMamber \cite{ReMamber}, and DETRIS \cite{DETRIS}. As shown in \cref{tab:zero_shot}, GeoLaV achieves 47.0 on Ref-Youtube-VOS and 31.6 on MeViS, improving upon SAMWISE by \textbf{+15.1} and \textbf{+5.2} $\mathcal{J}\&\mathcal{F}$, indicating notable generalization from image pretraining to unseen video domains.

\begin{table}[t]
\centering
\scriptsize
\setlength{\belowcaptionskip}{1mm}
\setlength{\tabcolsep}{2.0mm}
\renewcommand\arraystretch{1.2}
\definecolor{lightgray}{gray}{0.9}
\definecolor{lightblue}{RGB}{220, 240, 255}
\caption{Zero-shot performance when trained on images and evaluated on video benchmarks. All models are trained on RefCOCO/+/g \cite{RefCOCO,RefCOCOg} only.}
{
\begin{tabular}{l|ccc|ccc}
\hline
\multirow{2}{*}{Method} 
 & \multicolumn{3}{c|}{Ref-YouTube-VOS \cite{Ref-Youtube-VOS}} & \multicolumn{3}{c}{MeViS \cite{MeViS}} \\
\cline{2-7}
 & $\mathcal{J}\&\mathcal{F}$ & $\mathcal{J}$ & $\mathcal{F}$ & $\mathcal{J}\&\mathcal{F}$ & $\mathcal{J}$ & $\mathcal{F}$ \\
\hline
\rowcolor{lightgray} \multicolumn{7}{l}{\textit{\textbf{Image}-based Model}} \\
DMMI \cite{DMMI} \textcolor{gray}{\scriptsize[ICCV'23]} & 9.6 & 6.6 & 7.5 & 7.1 & 6.9 & 7.3 \\
ReMamber \cite{ReMamber} \textcolor{gray}{\scriptsize[ECCV'24]} & 35.1 & 35.3 & 35.0 & 26.5 & 24.6 & 28.4 \\
DETRIS \cite{DETRIS} \textcolor{gray}{\scriptsize[AAAI'25]} & 17.6 & 17.9 & 17.4 & 16.3 & 16.1 & 16.5 \\

\hline
\rowcolor{lightgray} \multicolumn{7}{l}{\textit{\textbf{Video}-based Model} (Trained on Images Only)} \\

LMPM \cite{MeViS} \textcolor{gray}{\scriptsize[ICCV'23]} & 7.1 & 7.0 & 7.1 & 11.8 & 11.6 & 12.0 \\
DsHmp \cite{DsHmp} \textcolor{gray}{\scriptsize[CVPR'24]} & 7.1 & 7.0 & 7.1 & 11.8 & 11.6 & 12.0 \\
LAVT \cite{LAVT} \textcolor{gray}{\scriptsize[TPAMI'25]} & 28.6 & 28.9 & 28.4 & 21.8 & 22.2 & 21.6 \\

SAMWISE \cite{SAMWISE} \textcolor{gray}{\scriptsize[CVPR'25]} & 31.9 & 31.2 & 32.6 & 26.4 & 24.6 & 28.1 \\
\rowcolor{lightblue} \textbf{GeoLaV (Ours)} & \textbf{47.0}& \textbf{46.9} & \textbf{47.2} & \textbf{31.6} & \textbf{29.6} & \textbf{33.6} \\
\hline
\end{tabular}}
\label{tab:zero_shot}
\end{table}

\noindent\textbf{Effectiveness of the Monocular Geometry Pretraining.}
As shown in \cref{tab:ablations}, introducing Monocular Geometry Pretraining (Stage~I) boosts the $\mathcal{J}\&\mathcal{F}$ score from 65.1 to 67.2. Compared with planar homography-based multi-frame augmentation, which yields only moderate gains, our depth-based warping strategy brings consistently larger improvements. This indicates that the performance gain does not merely stem from increased frame exposure, but from enforcing cross-view geometric consistency during pretraining. By synthesizing geometry-consistent novel views from single images, the encoder learns structure-aware representations that are more robust to viewpoint variations, providing a stronger initialization for subsequent video fine-tuning.

\noindent\textbf{Effectiveness of the Geometry-Aware Distillation.}
\cref{tab:ablations} shows that applying Geometry-Aware Distillation (Stage~II) improves $\mathcal{J}\&\mathcal{F}$ from 65.1 to 67.5. 
By aligning the video model with 3D-aware teacher features during fine-tuning, this stage injects structural and depth-aware priors that promote geometry-consistent representations across frames. 
Compared with generic feature supervision, such priors help maintain more stable object representations under viewpoint changes and motion. 
Moreover, combining Stage~II with Stage~I leads to substantially larger gains, suggesting that geometry-consistent pretraining provides a better initialization for absorbing 3D structural cues.

\begin{table}[t]
\centering
\scriptsize
\setlength{\belowcaptionskip}{1mm}
\setlength{\tabcolsep}{2.5mm}
\renewcommand\arraystretch{1.2}
\caption{Ablation study on the effectiveness of Monocular Geometry Pretraining (MGP) and Geometry-Aware Distillation (GAD). All models are first pretrained on image datasets and then fine-tuned on video benchmarks. Planar augmentation denotes homography-based 2D transformations (\textit{e.g.}, translation, scaling, rotation, cropping) that generate pseudo multi-frame inputs without geometric modeling.}
\begin{tabular}{l|c|c|ccc}
\hline
Setting 
& Stage I
& Stage II
& $\mathcal{J}\&\mathcal{F}$ & $\mathcal{J}$ & $\mathcal{F}$ \\
\hline
Vanilla Model 
& Base 
& Base
& 65.1 & 63.1 & 66.9 \\

+ Planar Augmentation 
& Planar 
& Base
& 66.5 & 64.5 & 67.6 \\

+ MGP (Ours)
& MGP 
& Base
& 67.2 & 66.1 & 68.0 \\

+ GAD (Ours)
& Base 
& GAD
& 67.5 & 66.2 & 68.9 \\

Full Model (Ours)
& MGP 
& GAD
& \textbf{70.5} & \textbf{69.1} & \textbf{71.8} \\
\hline
\end{tabular}
\label{tab:ablations}
\end{table}

\begin{figure}[t]
    \centering
    \setlength{\belowcaptionskip}{0mm}
   \setlength{\abovecaptionskip}{2mm}
    \begin{minipage}[t]{0.20\linewidth}
        \centering
        \includegraphics[width=\linewidth]{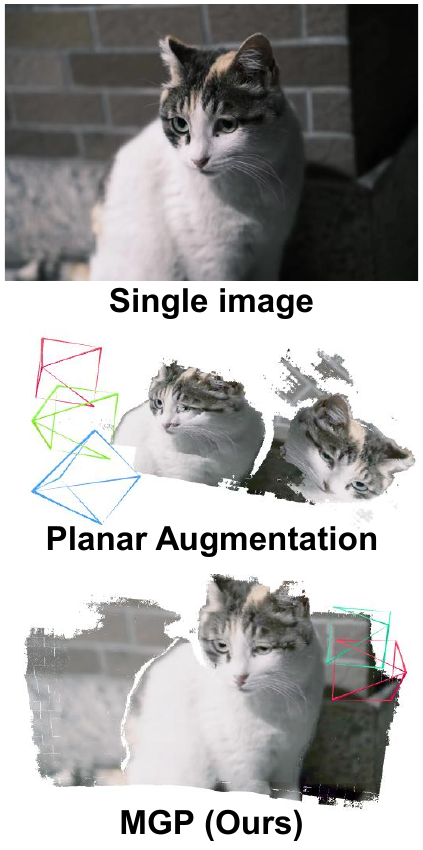}
        \captionof{figure}{Geometric consistency comparison: planar vs. ours.}
        \label{fig:geometry_compare}
    \end{minipage}
    \hfill
    \begin{minipage}[t]{0.78\linewidth}
        \centering
        \includegraphics[width=\linewidth]{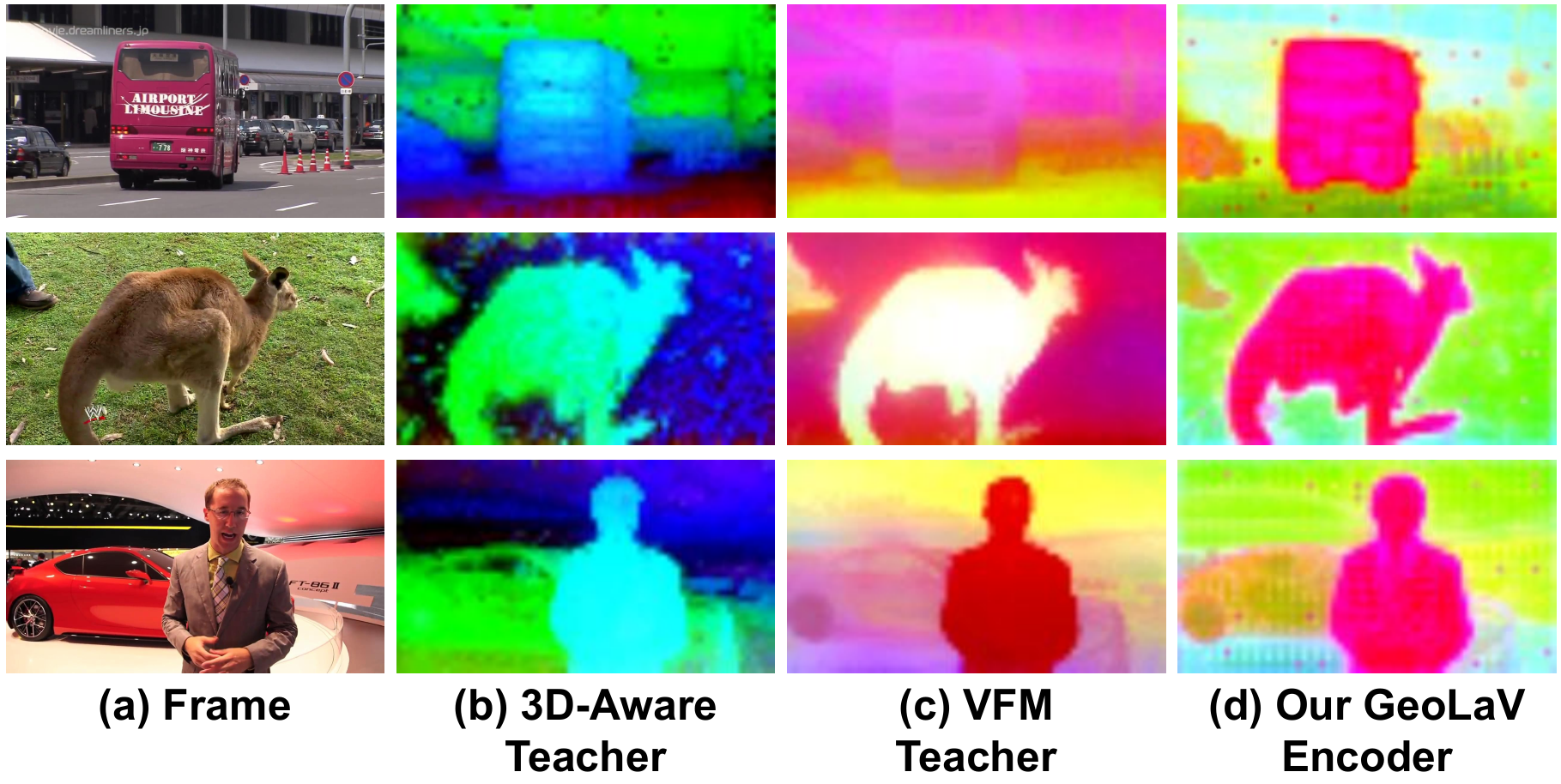}
        \captionof{figure}{Visualization of feature representations after geometry-aware distillation via PCA~\cite{pca}. GeoLaV integrates geometric and semantic cues from dual teachers, yielding more distinguishable object-level features.}
        \label{fig:feature}
    \end{minipage}
\end{figure}

\noindent\textbf{Geometric Accuracy Analysis.}
To validate that our synthesis preserves true 3D geometry, we compare it with homography-based planar transformations in \cref{fig:geometry_compare}. Planar augmentation applies 2D warping without modeling scene depth, often causing distorted shapes and incorrect occlusion under viewpoint changes. In contrast, our method reconstructs a 3D representation and reprojects it under continuous camera motion, preserving object geometry and depth ordering across views. This demonstrates that our synthesis maintains physically consistent 3D structure rather than performing purely 2D augmentation.

\noindent\textbf{Feature Visualization.}
Beyond geometric consistency at the image level, we further analyze how the learned representations encode structural and semantic information. To understand what GeoLaV learns from the two teachers, we visualize the encoder features extracted from the 3D-aware and VFM encoders in \cref{fig:feature}. The 3D-aware teacher provides depth and geometric cues, while the VFM teacher captures clear semantic boundaries and object-level understanding. After geometry-aware distillation, our encoder integrates both advantages, preserving geometric structure and maintaining sharp semantic edges. This shows that the dual-teacher design produces geometry-preserving and semantically consistent representations for precise video segmentation.

\section{Conclusion}
\label{sec:conclusion}

In this paper, we introduce \textbf{GeoLaV}, a two-stage framework that enhances text-driven video object segmentation through geometry-guided learning. Our GeoLaV addresses the limitations of prior 2D-based methods, which lack geometric consistency and spatial awareness. GeoLaV first employs monocular geometry pretraining that synthesizes novel views from single images, enabling geometry-consistent visual representation learning. Then, a geometry-aware distillation stage transfers 3D structural knowledge from general 3D priors, reinforcing spatiotemporal coherence and improving visual-language alignment. Extensive experiments on multiple benchmarks show that GeoLaV achieves state-of-the-art performance, validating the benefit of incorporating geometric reasoning into multimodal video understanding. Future work includes extending this framework toward open-vocabulary and large-scale 3D-aware multimodal segmentation.


\section*{Acknowledgements}
This work is supported by the National Natural Science
Foundation of China (62331006), the Fundamental Research Funds for the Central Universities, and the National Natural Science Foundation of China under Grant (625B2026).

%
%
\bibliographystyle{splncs04}
\bibliography{main}
\end{document}